\documentclass[letterpaper, 10 pt, conference]{ieeeconf}  

\IEEEoverridecommandlockouts                              

\usepackage{graphics} 
\usepackage{graphicx}
\usepackage{epsfig} 
\usepackage{mathptmx} 
\usepackage{times} 
\usepackage{amsmath} 
\usepackage{amssymb}  
\usepackage{bm}

\usepackage{array}
\newcolumntype{P}[1]{>{\centering\arraybackslash}p{#1}}

\title{\LARGE \bf
Advances in Hybrid Modular Climbing Robots: Design Principles and Refinement Strategies*
}

\author{Ryan Poon and Ian Hunter
\thanks{*The authors would like to thank Fonterra Co-operative Group Limited for sponsoring this work.}
\thanks{The authors are with the Bioinstrumentation Laboratory, Massachusetts Institute of Technology, Cambridge, MA 02139, United States of America (email: rpoon@mit.edu; ihunter@mit.edu)}
}

\begin{document}

\maketitle
\thispagestyle{empty}
\pagestyle{empty}

\begin{abstract}

This paper explores the design strategies for hybrid pole- or trunk-climbing robots, focusing on methods to inform design decisions and assess metrics such as adaptability and performance. A wheeled-grasping hybrid robot with modular, tendon-driven grasping arms and a wheeled drive system mounted on a turret was developed to climb columns of varying diameters. Here, the key innovation is the underactuated arms that can be adjusted to different column sizes by adding or removing modular linkages, though the robot also features capabilities like self-locking (the ability of the robot to stay on the column by friction without power), autonomous grasping, and rotation around the column axis. Mathematical models describe conditions for self-locking and vertical climbing. Experimental results demonstrate the robot's efficacy in climbing and self-locking, validating the proposed models and highlighting the potential for fully automated solutions in industrial applications. This work provides a comprehensive framework for evaluating and designing hybrid climbing robots, contributing to advancements in autonomous robotics for environments where climbing tall structures is critical.

\end{abstract}


\section{INTRODUCTION}

Climbing robots have a wide variety of applications over many industries involving vertical structures, ranging from monitoring road poles for cleaning to assessing damage on pipes to pruning tree branches. Automating these roles mitigates the dangers associated high heights and physically demanding tasks \cite{song_comprehensive_2022}. Decades of research have been committed towards developing these robots as well as reviewing, condensing, and classifying the diversity of the different design approaches taken to tackle the challenges each climbing environment introduces \cite{fang_advances_2023}.

In particular, climbing poles and trees---collectively called ``columns" from here onward---has been a common focus of robotics research in the past. Many past approaches involved grippers that took turns grasping the column while the body extended or compressed during a full gait cycle, such as 3DCLIMBR \cite{tavakoli2008}, Climbot \cite{guan2011}, Treebot \cite{lam2011}, CCRobot \cite{zheng2018}, and a variety of low-cost prototypes like \cite{norfaizal2015, li2015}. However, these robots faced a number of disadvantages, namely: slow climbing speed, restriction to a small range of column diameters for climbing, damage caused to the column, or poor payload capacity limited by design and grasping strength. 

Some researchers swapped the grasping mechanisms with wheels, sacrificing obstacle avoidance abilities and flexibility for climbing stability, electromechanical simplicity, and climbing speed. Many robots such as UT-PCR \cite{baghani2005} and others \cite{liu2021, franko2020} had successfully demonstrated wheeled vertical climbing of poles and trees, sometimes at the expense of climbing surface integrity \cite{ren2014}, \cite{shokripour2010}. Others like RETOV \cite{urdaneta2012} and \cite{gui2017} had the capability to rotate about the column axis, allowing more efficient column navigation. Furthermore, some wheeled robots could self-lock to stay on a column when all motors were disabled, which was ideal both for safety and to conserve battery energy for attached payloads. These robots---Pobot V2 \cite{fauroux2010} and UT-PCRII \cite{sadeghi2012}, for instance---exhibited this behavior by offsetting the center of mass away from the column axis and increasing friction. However, like the other aforementioned wheeled robots, their ring shaped body designs limited the range of column diameters that could be climbed, and several of these robots \cite{fischer2010}, \cite{kawaski2008} couldn't rotate about the column axis. These wheeled robots also had to be assembled around the column by a human before operation, which defeated the purpose of fully automating tasks that required climbing. 
%
\begin{table}
\vspace{0.1 in}
\caption{Climbing Robot Types: Advantages and Disadvantages}
\label{tab:proscons}
\centering
\begin{center}
\begin{tabular}{|P{14mm}|P{27mm}|P{27mm}|}
\hline
\textbf{Locomotion} & \textbf{Advantages} & \textbf{Disadvantages}\\
\hline
Wheeled & Continuous motion, simple control, high stability, high safety, high load capacity & Poor flexibility, poor obstacle avoidance ability, poor adaptability\\
\hline
Clamping & Stable, flexible, strong obstacle avoidance ability & Slow movement speed, complicated control, lower safety, lower load capacity\\
\hline
Bionic & Flexible, strong obstacle avoidance capability, often modular & Complicated control, poor load capacity, large force and torque requirements \\
\hline
Hybrid & Above advantages: simple control, stable, high load capacity, adaptive, modular & Higher system complexity, more DOFs, potentially poor obstacle avoidance \\
\hline
\end{tabular}
\end{center}
\end{table}
\begin{table}[ht]
\vspace{0.1 in}
\caption{Hybrid Climbing Robot Design Constraints}
\label{tab:constraints}
\centering
\begin{center}
\begin{tabular}{|P{22mm}|P{48mm}|}
\hline
\textbf{Constraint} & \textbf{Reasons} \\
\hline
Low Mass & Reduce drive power consumption and overall cost \\
\hline
Simple Build & Lower complexity generally means low risk of breakage \\
\hline
Non-Destructive Climbing & Climbing surface typically hard to puncture, surface damage undesirable \\
\hline
Self-Locking & Safety, can commit all power to payload \\
\hline
Automated Grasping & Removes need for human assembly before climbing \\
\hline
Adaptive to Column Diameter & Broadens variety of tasks, allows frequent climbing, limits max column diameter \\
\hline
\end{tabular}
\end{center}
\end{table}

Summaries of the pros and cons of wheeled and grasping climbing robots \cite{song_comprehensive_2022, zhang2021} are compiled in Table \ref{tab:proscons}. Table \ref{tab:proscons} also includes bionic robots like Climbot \cite{guan2011} and HyDRAS \cite{goldman_considerations_2008}, which can be perceived as a subset of clamping robots with similar advantages and disadvantages but also are often uniquely modular in design and adaptable to column diameters. Combining the advantages of these three main groups is one of the next exploratory steps in climbing robot advancement. 

To start, past research \cite{song_comprehensive_2022, fang_advances_2023, zhang2021} has provided overviews of climbing robot research from which several commonly desired features across many projects can be formulated as desired design constraints for future climbing robots (Table \ref{tab:constraints}). In this study, a hybrid climbing robot is characterized as one that satisfies all the design constraints in Table \ref{tab:constraints} by adopting the key defining features of the locomotion modes in Table \ref{tab:proscons}: a wheeled drive (wheeled climbers), grasping arms (clamping), and a modular system (bionic). To the authors' knowledge, only two robots have used both grasping arms and wheels for climbing \cite{sebastian2015, liu2022}. However, neither robot has the adaptability that comes with the truly modular systems that bionic robots possess. Units can be added to their grasping arms to accommodate different columns, but their designs constrain these units to be electromechanically complex, nonuniform, and bulky. 

\begin{figure}
    \centering
    \includegraphics[width = 0.99 \linewidth]{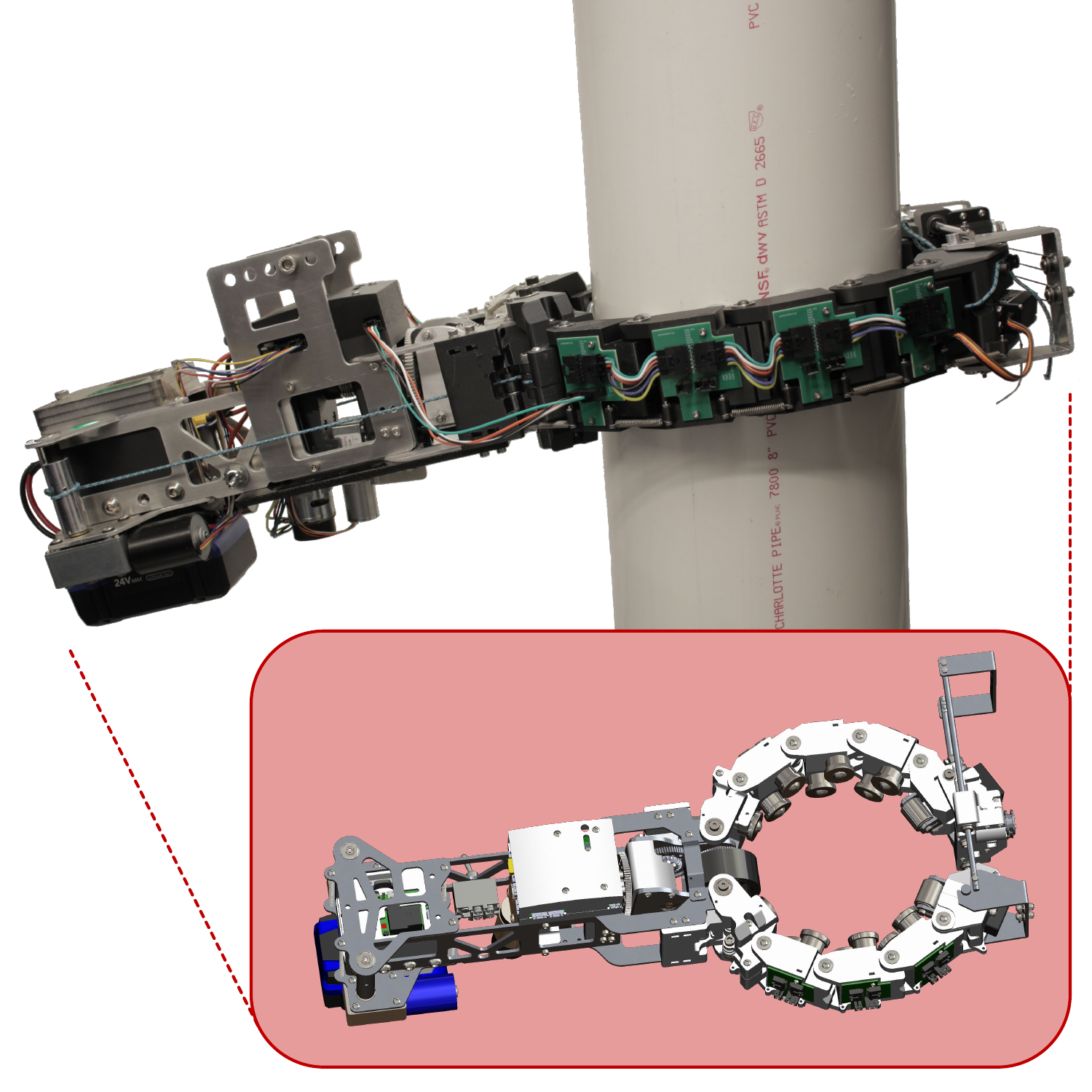}
    \caption{The hybrid climbing robot CLIMR: Cabled Limb Interlocking Modular Robot.}
    \label{fig:robot}
\end{figure}
The focus of this work is twofold. First, this work presents CLIMR, a hybrid wheeled robot with two modular, tendon-driven grasping arms, constructed in attempt to maximize adaptability to different column diameters while also featuring self-locking, autonomous grasping, and rotation about column axis (Fig. \ref{fig:robot}). From here, this work then introduces general design strategies for hybrid climbing robots, devising methods to evaluate future work and guide design decisions via metrics like adaptability and performance. In this way, this study on hybrid climbing robots takes one step closer towards a fully automated solution in the various industries they would be useful in. To summarize, the contributions of this work are as follows: 

\begin{enumerate}
    \item A novel modular wheeled-grasping robot that exploits the inherently higher movement speed and control simplicity of wheeled robots while mitigating the effects of the disadvantages listed in Table \ref{tab:proscons}. 
    \item Metrics and methods to evaluate the effectiveness of hybrid climbing robots and inform design choices.   
    \item Experimental results that validate the presented models and methods while demonstrating the viability of the presented design. 
\end{enumerate}
%

\section{CLIMR SYSTEM DESIGN}
\label{sec:design}

The hybrid climbing robot presented in this work consists of two modular grasping arms and a wheeled drive system, with the main body on a cantilevered tail to offset the center of mass away from the column. The drive wheel on the body is mounted on a turret, enabling both vertical climbing and rotation about the column. 

\subsection{Modular Arm Design}

Each grasping arm comprises of multiple modular linkages. A single UHMWPE cable passes through all the linkages, acting as the driving ``tendon" in this underactuated system. This tendon-driven design has several benefits, as it 1) raises the maximum climbable column diameter by reducing modular unit mass, as will be seen in Section \ref{sec:designmethod}, 2) allows the tendon-driving worm gear motors to be mounted on the cantilever tail to assist with self-locking, and 3) allows the addition of more links for a longer arm by just increasing tendon length without increasing electromechanical complexity. 

\begin{figure}
\vspace{0.1 in}
    \centering
    \includegraphics[width = 0.9 \linewidth]{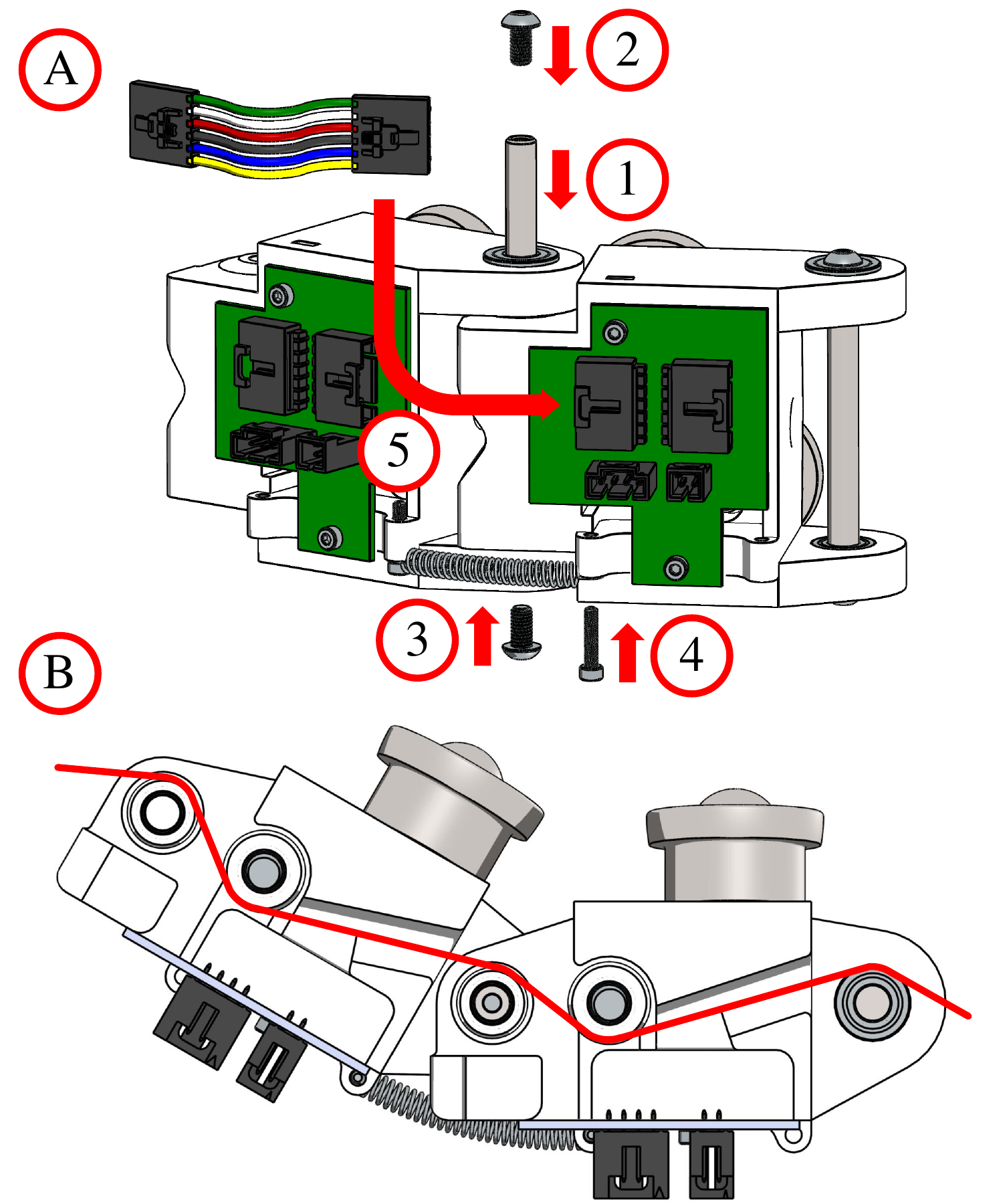}
    \caption{Adding a link to increase length of modular arm. (A) shows the process of adding a link: insert rotary shaft (1), add screws (2 and 3), connect the sockets on the PCB (4), and connect the antagonist spring (5). (B) shows the tendon path.}
    \label{fig:linkattachment}
\end{figure}
Connecting each link is a rotational pin joint, and the process of attaching or removing linkages for different column diameters can be seen in Fig. \ref{fig:linkattachment}. To contact the column and improve grasping stability, each link has two steel ball transfer bearings. These minimize sliding friction that resists climbing forces, and a minimum of two ball transfers per link are required to prevent the link from slipping out of its intended orientation when the tendon is tightened during grasping. The tendon is guided through each link via small idler pulleys to reduce the friction along the tendon path and facilitate the restorative backwards movement of the link driven by low-stiffness antagonist springs. Orienting the pulleys as seen in Fig. \ref{fig:linkattachment} allows for sequential bending, where each link bends only when the previous one has bent fully. Serrated flanged screws on the end links fix the tendons.


The designs of the first and last links on the arms are different than that of the standard modular links. Because the end links experience the most force due to the moment generated by the cantilever tail, the end links have a clutch roller bearings mounted on stages rotated by BLS4060MED brushless servos. This is to reduce friction for climbing but also to assist with self-locking via the unidirectional nature of the clutch roller bearings. 

\subsection{Latch Design}
\label{subsec:designlatch}

On one end link is an automated latch that hooks onto a pin on the opposite end link and draws the arms together for more secure grasping before climbing. This latch comprises of one rotary and one prismatic joint. When the arm links are fully contacting the column, continuing to pull the tendon causes the latch to rotate about its rotary joint until the latch reaches the pin on the opposite end link. A 15 mm micro worm gear motor from Precision Mini Drives then draws the latch in along the linear shafts via a 1 mm diameter UHMWPE cable and compound pulley system until the hook catches the rod. Because tendon drives can only pull in one direction, a constant force spring provides the restoring force so the hook latch can return to its original position. 


\subsection{Body and Tail Design}

\begin{figure}
\vspace{0.1 in}
    \centering
    \includegraphics[width = 0.99 \linewidth]{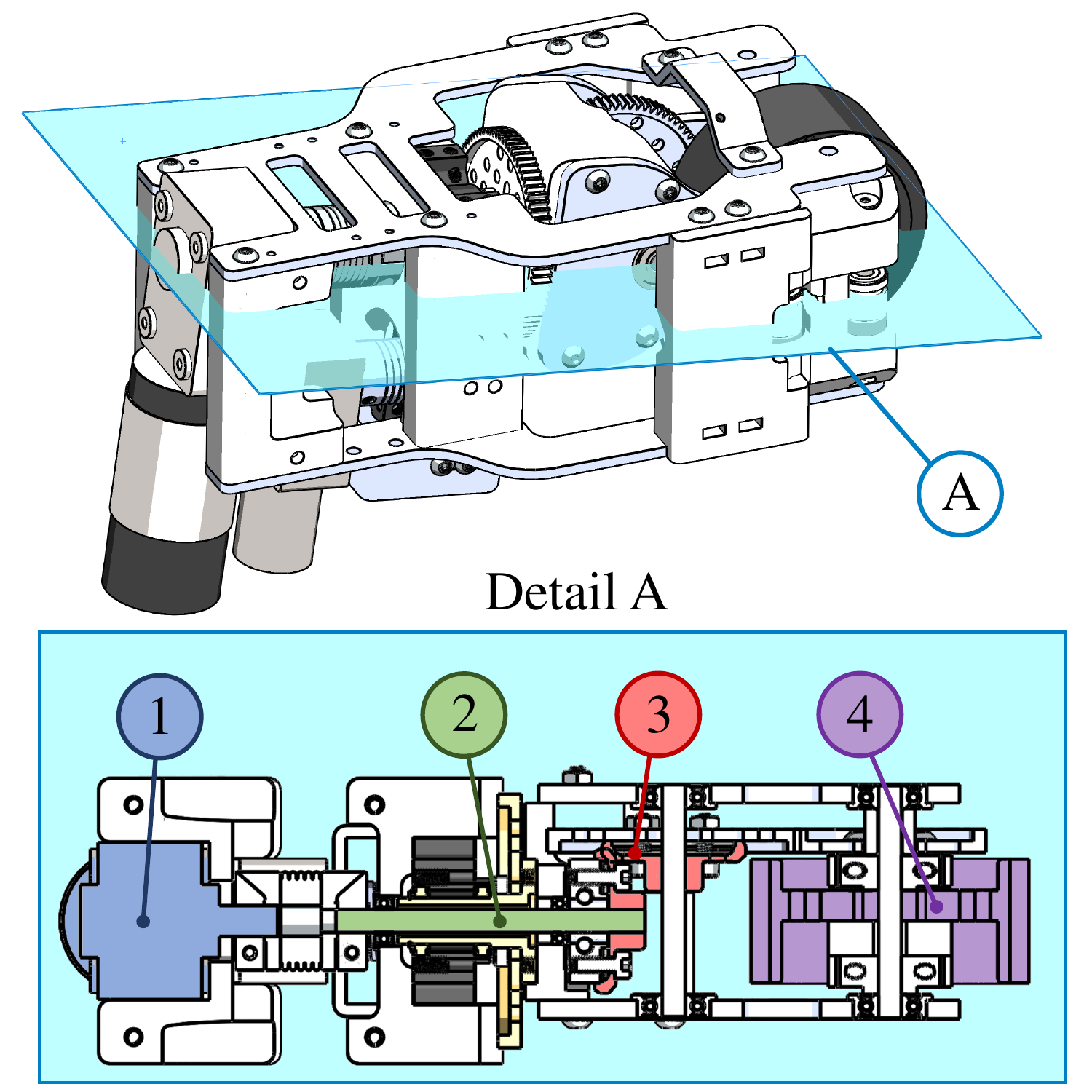}
    \caption{The body of the robot and a cross section of wheeled turret drive system. In Detail A, the motor (1) drives the through-hole drive shaft (2) and the bevel gears (3) to rotate the drive wheel (4).}
    \label{fig:drivesystem}
\end{figure}
The main body houses the drive system. A 5840WG-555PM E-S Motors brushed motor with a 280:1 worm gearbox drives the 30A rubber wheel via a through-hole driveshaft and a bevel gear transmission, which is housed in a turret. A 4632WG-2430BL E-S Motors brushless motor with a 600:1 worm gearbox rotates the turret (Fig. \ref{fig:drivesystem}). In this way, both vertical climbing and rotation about the column is achievable. The drive wheel has to be in the workspace plane of the arms---otherwise, wheel slip would occur due to the extra degrees of freedom in the modular arms.    

The cantilever tail shifts the center of mass away from the axis of the column, creating a moment arm that increases the normal force on the drive wheel and facilitates self-locking. It houses the tendon driving motors, the robot's computer, and a Kobalt tool battery to reduce the amount of ``non-functional" weight required to achieve self-locking behavior. However, the design of the tail incorporates mounting points for modular tail weights to shift the center of mass further rearward if needed. 

\subsection{Electronics and Programming Infrastructure}

The main computer is a Raspberry Pi 4B with Ubuntu and ROS2. All commands to the motors are sent via CANBus to a custom PCB on the body that has a Teensy 4.1 microcontroller and several breakout boards. The Teensy is connected to a G2 18v25 Pololu motor driver for the drive wheel, two INA169 current sensors for the tendon motors, and a DRV8876 Pololu brushed motor driver for the latch. PWM commands are sent directly to the brushless tendon motors, the turret motor, and the servos. The system is powered with 24 V, which is stepped down via buck converters to 5 V for the Raspberry Pi and Teensy and 12 V for the servos and latch micro motor. A relay and manual switch on the tail are used as remote and manual emergency-stops. 

Standard PID position controllers are implemented for most of the actuators. However, the derivative components of the drive motor's PI velocity controller and the tendon motors' PI torque controllers are removed due to noise in the feedback readings. The micro worm gear motor for the latch is run until a current spike is detected, indicating that the latch has securely hooked onto the opposite end link. 


\section{DESIGN METHOD}
\label{sec:designmethod}

\subsection{Definitions}

The development of CLIMR sheds light on the relationship between design choices and its limitations, which can generalized into a methodology for future development of hybrid climbing robots to mitigate those limitations through calculated design decisions. When designed as modular systems, these hybrid robots can adapt to columns of different sizes by inserting additional links into their structure. As with any climbing robot, the maximum column diameter will be limited by the drive motor torque and overall weight. However, for hybrid robots, strategic design choices can raise this upper bound without a complete redesign.

The core of the hybrid climbing robot is its modularity and its ability to adapt to different column diameters $d_c$. Generally, this can be expressed as a mathematical conditional statement,

\begin{equation}
\label{eq:generalnumlinks}
n' = 
\begin{cases}
n+1 & \text{if } f(n,d_c) < c \\
n & \text{if } f(n,d_c) \geq c
\end{cases},
\end{equation}
where $n'$ is the next value of $n$, and $n$ is the number of modular units that is dependent on some function $f$ and some user-defined constant limit $c$. The function $f$ could be based on some geometric or statics constraint, such as the location of the robot's center of mass or the distance between a component and the column. 

The value $n$ determines the distribution of mass and thus dictates the conditions for self-locking and vertical climbing. For self-locking, the condition often takes the form of a lower bound (\cite{fauroux2010} is one such example),

\begin{equation}
\label{eq:generalselflock}
d_\text{COM} < g(n,d_c),
\end{equation}
where $d_\text{COM}$ describes the distance between the center of mass and the axis of the column. The function $g(n,d_c)$ is typically derived from static force balance equations and depends on some dimensions of the robot. Similarly, the conditions for successful vertical climbing depend on the force $F_\text{dr}$ tangential to the drive wheel,

\begin{equation}
\label{eq:generalvertclimb}
h_1(a_d) \leq F_\text{dr} \leq h_2(n,d_c),
\end{equation}
where $F_\text{dr}$ is lower bounded by function $h_1$ that depends on the desired acceleration $a_d$ and upper bounded by function $h_2$ that describes the friction between the drive and the climbing surface. If this $F_\text{dr}$ that is generated by the drive motor and transmission exceeds $h_2$, the wheel slips. 

\subsection{Methodology}
\label{subsec:methodology}

With the definition of hybrid climbing robots and their design constraints established, the following methodology determines the maximum diameter of the column that can be climbed and provides metrics to evaluate the resulting robot design. As such, it gives guidance on how to tune certain design factors to optimize climbing adaptability. Qualitatively, Equations \ref{eq:generalnumlinks}, \ref{eq:generalselflock}, and \ref{eq:generalvertclimb} can be consolidated into the following iterative procedure: 

\begin{enumerate}
    \item Add a modular unit $n' = n+1$ when $f(n,d_c)$ falls below a constant $c$ according to Equation \ref{eq:generalnumlinks}.
    \item Check that self-locking condition is satisfied according to Equation \ref{eq:generalselflock}. 
    \item Check that vertical climbing conditions are satisfied according to Equation \ref{eq:generalvertclimb}.
\end{enumerate}
The maximum column diameter $d_c$ has been reached when the conditions in Equations \ref{eq:generalselflock} and \ref{eq:generalvertclimb} are not met. This value $d_c$ gives a metric by which one can evaluate a robot design for a given motor torque and desired acceleration. From this, it can be seen that the mass of the modular units is critical to the hybrid robot adaptability. Modular units with added weight from motors or complex mechanisms reduce the maximum climbable column diameter, which is why CLIMR's modular arm system is an underactuated tendon driven mechanism. It's advantageous to make modular unit mass as low as possible so the total mass is not increased beyond the capabilities of the drive system or so the center of mass is not shifted too far and the property of self-locking is lost. This insight is an integral part in optimizing for maximum column diameter and is further explored by modeling CLIMR in Section \ref{sec:modeling}.

\section{MODELING AND EQUATIONS}
\label{sec:modeling}

To determine the upper limit of climbable column diameter, the modeling of CLIMR was done to attain expressions for functions $f(n,d_c)$, $g(n,d_c)$, $h_1(a_d)$, and $h_2(n,d_c)$ in Equations \ref{eq:generalnumlinks}, \ref{eq:generalselflock}, and \ref{eq:generalvertclimb}. From there, executing the steps in Subsection \ref{subsec:methodology} would give an idea of the resulting adaptability and performance. Modeling of other features like turret rotation is included for completeness.

\subsection{Determining Linkage Number}
\label{subsec:numlinks}

\begin{figure}
    \centering
    \includegraphics[width = 0.9 \linewidth]{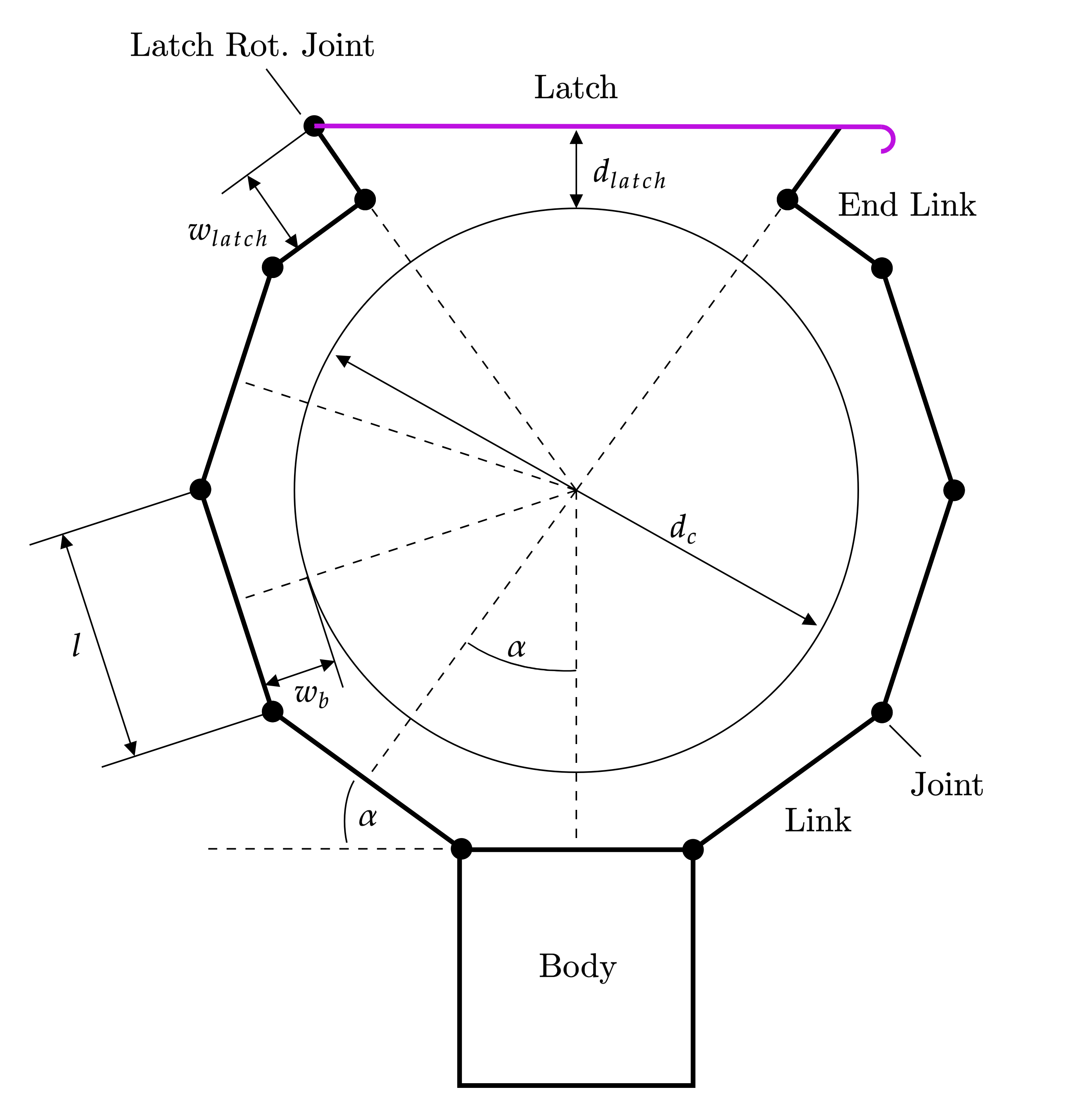}
    \caption{Line diagram with the important dimensions to determine the relationship between number of links and column diameter.}
    \label{fig:linediagramnumlinks}
\end{figure}

The column diameter $d_c$ for a given number of links $n$ per arm is limited by the distance between the latch and the column $d_\text{latch}$ (where $d_\text{latch}=c$ from Equation \ref{eq:generalnumlinks}):

\begin{equation}
\label{eq:dlatch}
\begin{split}
d_\text{latch} = & \sum_{k=1}^{n-1}l\text{sin}(k\alpha) + \frac{l}{2}\text{sin}(n\alpha) 
\\
& + w_\text{latch}\text{sin}\left(n\alpha-\frac{\pi}{2}\right) - w_b - d_c.
\end{split}
\end{equation}
This assumes that the robot arms as a whole do not deform when deployed and hanging on a column. Here, $\alpha$ is the angle each link forms with respect to its neighbors, $l$ is the link length, $w_\text{latch}$ is the distance between the middle of the end link to the rotational joint of the latch holder, and $w_b$ is the distance between the middle of a link to its contact point on the column (Fig. \ref{fig:linediagramnumlinks}). In reality, $d_\text{latch}$ should be larger than zero to avoid collision of the latch against the column. The value $\alpha$ is given as

\begin{equation}
\label{eq:alpha}
\alpha = 2\text{tan}^{-1}\left(\frac{l}{2w_b+d_c}\right). 
\end{equation}

\subsection{Self-Locking and Driving}
\label{subsec:selflocking}

\begin{figure}[t]
    \centering
    \includegraphics[width = 0.99 \linewidth]{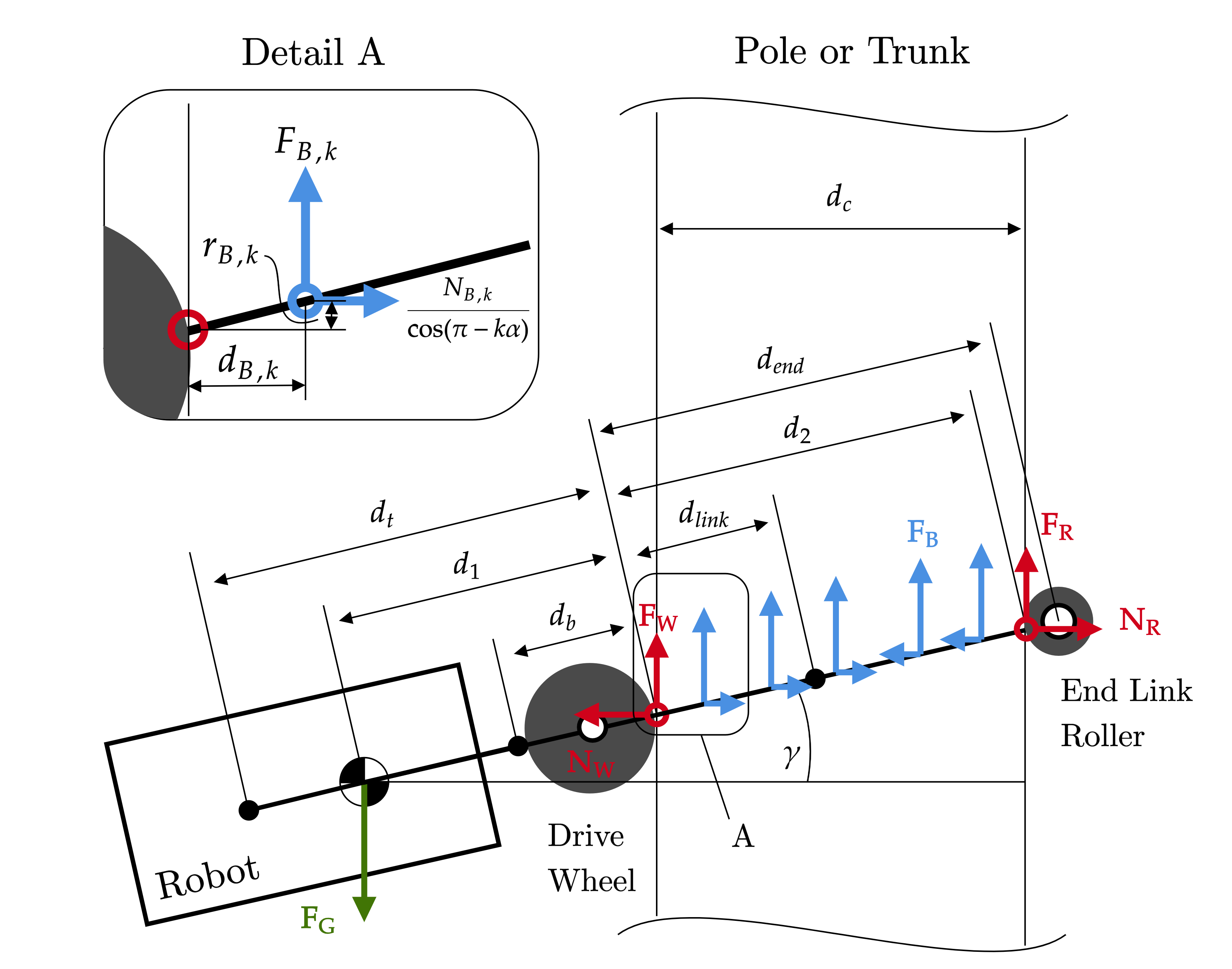}
    \caption{Line diagram with the important forces, dimensions, and center of mass locations to determine criteria for self-locking.}
    \label{fig:linediagramselflocking}
\end{figure}

To ensure that the robot can self-lock, the gravitational force $F_G$ must satisfy the inequality condition

\begin{equation}
\label{eq:equilibrium}
F_G \leq F_W + 2(F_B + F_R),
\end{equation}
where $F_W$, $F_B$, and $F_R$ are the forces of friction of the wheel, ball transfer bearings, and link roller against the climbing surface (Fig. \ref{fig:linediagramselflocking}). If it is assumed that $F_B=0$, the torques about the wheel-column contact point are

\begin{equation}
\label{eq:sumoftorques}
\begin{split}
0 = & F_Gd_1\cos(\gamma)+2F_Rd_2\cos(\gamma) \\
& -2N_Rd_2\sin(\gamma)\cos(\pi-n\alpha),
\end{split}
\end{equation}
where $\gamma$ is the angle the robot makes with the horizontal plane, $d_1$ is the distance from wheel contact to the center of mass, $d_2$ is the distance from wheel contact to end roller contact, and $N_R$ is the normal force on each roller (Fig. \ref{fig:linediagramselflocking}). Using Equations \ref{eq:equilibrium} and \ref{eq:sumoftorques} as well as the notion that $N_W=2N_R\cos(\pi-n\alpha)$ gives the condition for self locking,

\begin{equation}
\label{eq:conditionselflocking}
d_1 \geq \left|\frac{d_2(\tan(\gamma)\cos(\pi-n\alpha)-\mu_R)}{\mu_W\cos(\pi-n\alpha)+\mu_R}\right|,
\end{equation}
where $\mu_R$ is the friction coefficient between the end link rollers and the column ($F_R=\mu_RN_R$), and $\mu_W$ is the friction coefficient between the drive wheel and the column ($F_W=\mu_WN_W$). Equation \ref{eq:conditionselflocking} is analogous to Equation \ref{eq:generalselflock}, as the difference between $d_\text{COM}$ and $d_1$ is simply half of the constant $d_c$, and $g(n,d_c)$ can be attained easily. Though the robot is designed such that $d_2=d_c$, the robot arms are expected to deform due to the high number of degrees of freedom. It is physically impossible for $\gamma=0$ because there will always be deformation.



For climbing, assuming no ball transfer bearing friction and drive wheel inertia, the motor torque $\tau_\text{m}$ required to move the robot with desired acceleration $a_d$ is

\begin{equation}
\label{eq:motortorque}
\tau_\text{m} = \frac{r_w (ma_d + mg)}{G_\text{dr}},
\end{equation}
where $r_w$ is the radius of the drive wheel, $m$ is the mass of the robot, $g$ is the acceleration due to gravity, and $G_\text{dr}$ is the gear ratio of the drive transmission. Equation \ref{eq:motortorque} only holds true if the force generated by the drive $F_\text{dr} = \frac{G_\text{dr}\tau_\text{m}}{r_w}$ is less than the frictional force $F_W$ between the wheel and the climbing surface. Mathematically, from Equation \ref{eq:sumoftorques} and the force balance equation relating $N_W$ to $N_R$ (Fig. \ref{fig:linediagramselflocking}), 

\begin{equation}
\label{eq:noslipcondition}
F_\text{dr} \leq |F_W| = \left|\frac{mg d_1\mu_W\cos(\pi-n\alpha) }{d_2 \tan(\gamma)\cos(\pi-n\alpha) - \mu_Rd_2}\right|
\end{equation}
In this way, Equations \ref{eq:motortorque} and \ref{eq:noslipcondition} provides the bounds for $F_\text{dr}$ as described in Equation \ref{eq:generalvertclimb}.

From a practical sense, though, the friction of the ball transfer bearings does play a role in both self-locking and driving. If the ball transfer friction $F_B$ is not trivial due to the forces generated by the counterweight tail, the force balance in Equation \ref{eq:sumoftorques} must be adjusted to include it for self-locking. For driving, balancing the vertical forces in Fig. \ref{fig:linediagramselflocking} shows that $F_W$ in Equation \ref{eq:noslipcondition} can also be expressed as 

\begin{equation}
\label{eq:FWupperbound}
F_W = F_G-2(F_R+F_B).
\end{equation}
However, $F_B$ can be described as the sum $\sum_{k=1}^{n-1}F_{B,k}$, where $F_{B,k}$ is the friction force of the ball transfer of link $k$ (see Detail A of Fig. \ref{fig:linediagramselflocking}). Hence, for both self-locking and driving, getting an analytical expression of $F_B$ for Equations \ref{eq:sumoftorques} and \ref{eq:noslipcondition} is very challenging due to the nonuniform distribution of $F_B$ across the arms and the lack of ball transfer bearing friction characterization data. 

Regardless, these models taking into account $F_B$ provide useful insights for designing the robot to allow self-locking without compromising climbing. Firstly, it can be seen that Equation \ref{eq:conditionselflocking} is clearly a upper bound. This means that $d_1$ can be lower than what is proposed in Equation \ref{eq:conditionselflocking} if self-locking is the only concern. However, Equation \ref{eq:FWupperbound} shows that $F_B$ can get so large that it causes wheel slip. As the value for $d_1$ increases, $F_B$ also increases. Thus, successfully designing this climbing robot hinges on the balance between self-locking and climbing ability---making sure $d_1$ from Equation \ref{eq:conditionselflocking} is large enough for self-locking but not so large that $F_B$ inhibits climbing. 

%


\subsection{Torque for Turret Rotation}

\begin{figure}[t]
    \centering
    \includegraphics[width = 0.99 \linewidth]{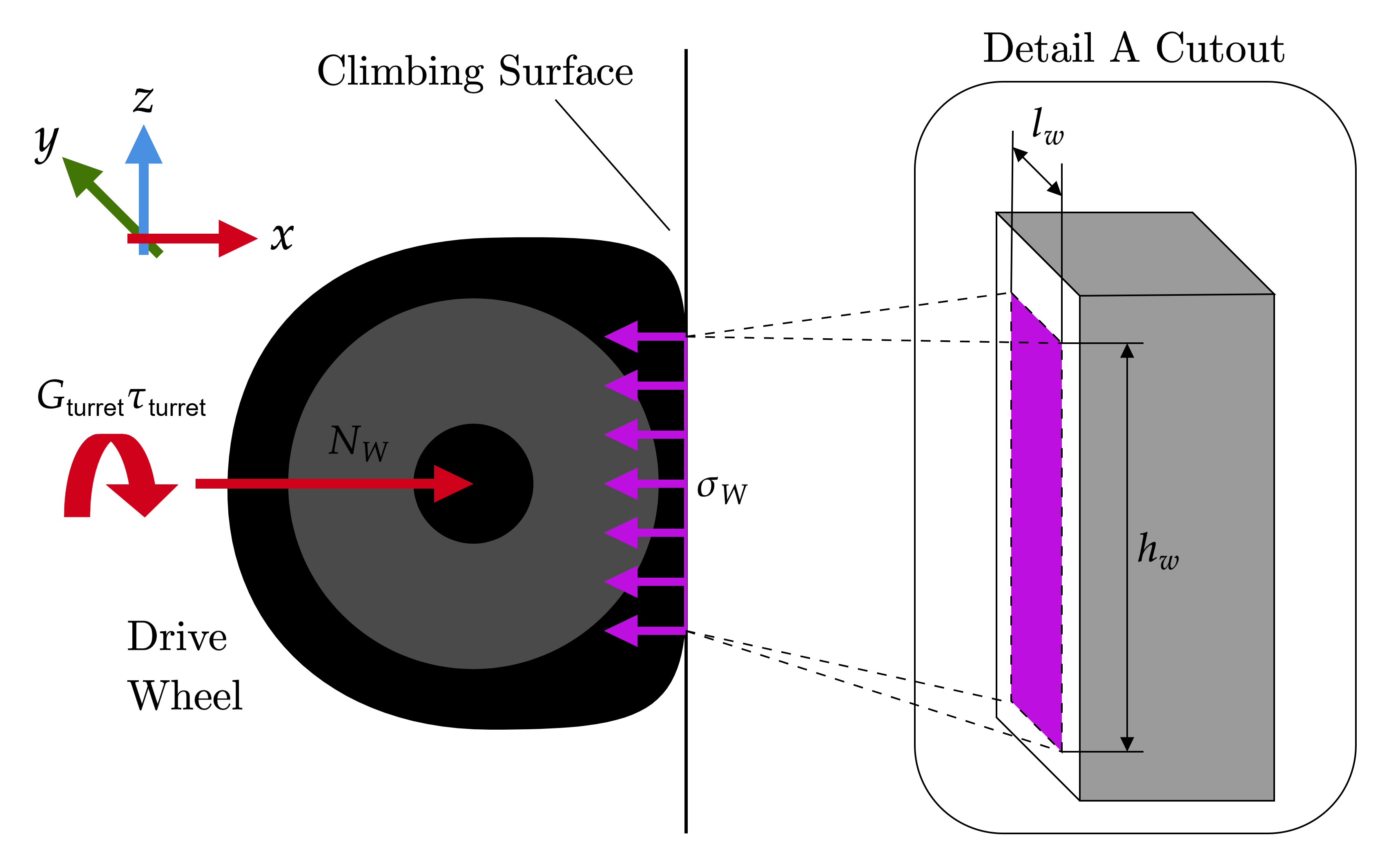}
    \caption{Line diagram with variables to calculate necessary turret torque due to wheel deformation.}
    \label{fig:linediagramturrettorque}
\end{figure}

The turret rotation motor must generate enough torque to overcome the frictional contact area created by the deforming rubber drive wheel. The normal force on the wheel $N_W$ can be deduced by $F_W =\mu_W N_W$ and Equation \ref{eq:noslipcondition}, 

\begin{equation}
\label{eq:NW}
N_W = \frac{mg d_1\cos(\pi-n\alpha)}{d_2 \tan(\gamma)\cos(\pi-n\alpha) - \mu_Rd_2}.
\end{equation}
From here, two assumptions are made. First, the contact interface with the climbing surface is simplified to be flat as opposed to concave (Detail A in Fig. \ref{fig:linediagramturrettorque}). Second, the normal force $N_W$ is assumed to be uniformly distributed across the area, where the normal force per unit area $\sigma_W$ can be described as

\begin{equation}
\label{eq:sigmaW}
\sigma_W = \frac{N_W}{l_wh_w} \text{ , } h_w = \frac{N_W}{E_\text{rubber}l_w},
\end{equation}
where $l_w$ is the length of the wheel, $h_w$ is the height of the compressed contact area (Fig. \ref{fig:linediagramturrettorque}), and $E_\text{rubber}$ is the rubber modulus of elasticity. Hence, the amount of motor torque $\tau_\text{turret}$ needed to overcome friction can be found by integrating across the contact area,

\begin{equation}
\label{eq:turrettorque}
\tau_\text{turret} = \int_{-\frac{h_w}{2}}^{\frac{h_w}{2}} \int_{-\frac{l_w}{2}}^{\frac{l_w}{2}} \frac{\mu_W\sigma_W}{G_\text{turret}}\sqrt{y^2 + z^2} dy\text{ }dz,
\end{equation}
where $G_\text{turret}$ is the gear reduction ratio of the transmission between the motor and the turret, and the $y$ and $z$ axes are according to the axes in Fig. \ref{fig:linediagramturrettorque}.

\subsection{Relationship between Linkage Number and Tail Weight}
\label{subsec:numlinklimitation}

As explained in Section \ref{sec:designmethod}, the upper limit to the maximum climbable column diameter is found when the total mass, which increases as new modular units are added, is too much for the drive motor. Framed around the proposed steps in Subsection \ref{subsec:methodology}, the following methodology is conducted until the motor torque $\tau_m$ is insufficient for climbing.

\begin{enumerate}
    \item Add a link $n = n+1$ when $d_\text{latch}$ gets below a minimum value according to Equation \ref{eq:dlatch}.
    \item Calculate $d_1$. 
    \item Check if Equation \ref{eq:noslipcondition} holds true. If not, add more weight $m_{tw}$ to the tail to increase $m_t$.
    \item Check if $\tau_m$ is enough to satisfy Equation \ref{eq:motortorque} for a desired acceleration $a_d$.
\end{enumerate}
The actual expression of $d_1$ (Fig. \ref{fig:linediagramselflocking}) for this robot is

\begin{equation}
\label{eq:comlocation}
d_1 = \left|\frac{2(n-1)m_\text{link}d_\text{link} + m_\text{end}d_\text{end} - m_bd_b - m_td_t}{m_b + m_t + 2(n-1)m_\text{link} + m_\text{end}}\right|,
\end{equation}
where $m_b$, $m_t$, $m_\text{link}$, and $m_\text{end}$ are respectively the masses of the body, tail, links, and end links with the latch mechanism. The dimensions $d_b$, $d_t$, $d_\text{link}$, and $d_\text{end}$ are the locations of each component's center of mass with respect to the wheel and climbing surface interface (Fig. \ref{fig:linediagramselflocking}). Both $d_\text{link}$ and $d_\text{end}$ change with link number,

\begin{equation}
\label{eq:dlink}
d_\text{link} \approx \frac{l}{2}\sum_{k=1}^{n-1}\text{sin}(k\alpha)-r_w,
\end{equation}
\begin{equation}
\label{eq:dend}
\begin{split}
d_\text{end} \approx & \sum_{k=1}^{n-1}l\text{sin}(k\alpha) + \frac{l}{2}\text{sin}(n\alpha) 
\\
& + w_\text{latch}\text{sin}\left(n\alpha-\frac{\pi}{2}\right) - r_w.
\end{split}
\end{equation}

\section{SIMULATIONS}
\label{sec:simulations}

To determine $d_1$, calculations were based on the parameters in Table \ref{tab:parametersstatics}. The desired acceleration $a_d$ was selected to design a robot that could achieve a climbing speed of 100 mm/s in less than 3 seconds, which is comparable to past work such as \cite{liu2021, gui2017, fauroux2010}. PVC pipe was chosen for testing due to its lower coefficient of friction with rubber. If the robot climbs a PVC pipe, it's likely to be able to climb surfaces of other materials, such as wood or steel. Using Equation \ref{eq:conditionselflocking}, $d_1$ was made to be $\geq$120 mm for the experimental tests. This ensured that $F_G$ in Equation \ref{eq:FWupperbound} was sufficient enough to prevent wheel slippage so CLIMR performed well in case of mass distribution discrepancies between CAD and reality. 

\begin{table}
\vspace{0.1 in}
\caption{Parameters for Simulation and Calculations}
\label{tab:parametersstatics}
    \centering
    \begin{tabular}{|c|c|c|c|}
        \hline
        \textbf{$m_b$ [kg]} & \textbf{$m_t$ [kg]} & \textbf{$m_\text{link}$ [kg]} & \textbf{$m_\text{end}$ [kg]} \\
        \hline
        1.873 & 3.311 & 0.221 & 0.754 \\
        \hline
        \hline
        \textbf{$m_{tw}$ [kg]} & \textbf{$d_b$ [mm]} & \textbf{$d_t$ [mm]} & \textbf{$\mu_W$} \\
        \hline
        0.261 & 154.6 & 329.3 & 0.7 \\
        \hline
        \hline
        \textbf{$\mu_R$} & \textbf{$\gamma$ [rad]} & \textbf{$l$ [mm]} & \textbf{$w_\text{latch}$ [mm]} \\
        \hline
        0.5 & 0.175 & 65 & 11 \\
        \hline
        \hline
        \textbf{$r_w$ [mm]} & \textbf{$a_d$ [rad/s$^2$]} & \textbf{$G_\text{dr}$} & \textbf{$G_\text{turret}$} \\
        \hline
        36 & 1 & 0.75 & 8/3 \\
        \hline
        \hline
        \textbf{$E_\text{rubber}$ [MPa]} & \textbf{$l_w$ [mm]} & \textbf{$g$ [m/s]} &  \\
        \hline
        0.7 & 32 & 9.81 & \\
        \hline
    \end{tabular}
\end{table}

\begin{table}
\caption{Motor/Gearbox Specifications (Equations \ref{eq:motortorque}, \ref{eq:noslipcondition}, and \ref{eq:turrettorque})}
\label{tab:motorselect}
    \centering
    \begin{tabular}{|c|P{20mm}|P{18mm}|P{18mm}|}
        \hline
        \textbf{Motor} & \textbf{Required Torque (N$\cdot$m)} & \textbf{Rated Motor Torque (N$\cdot$m)} & \textbf{Rated Motor Speed (RPM)} \\
        \hline
        Drive & 4.057 & 6.865 & 41 \\
        \hline
        Turret & 0.829 & 1.863 & 7 \\
        \hline
    \end{tabular}
\end{table}

\begin{figure}
    \centering
    \includegraphics[width = 0.99 \linewidth]{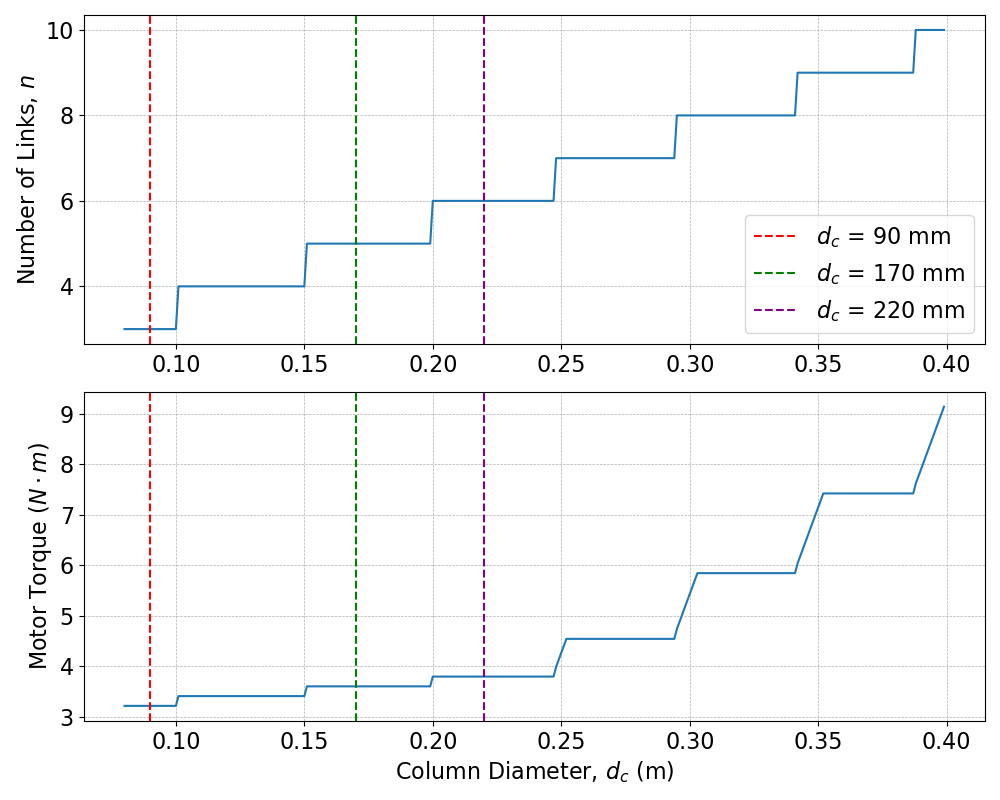}
    \caption{Torque required for added links and tail weights as column diameter increases. The red, green, and purple dashed vertical lines indicate the diameters of the columns used for testing.}
    \label{fig:numlinklimit}
\end{figure}
To evaluate the design, a depiction of the relationship between the column diameter $d_c$ and the required motor torque $\tau_m$ is found using the process at the end of Subsection \ref{subsec:numlinklimitation} and the parameters in Table \ref{tab:parametersstatics} (Fig. \ref{fig:numlinklimit}). Motors and gearboxes were selected according to torque values calculated with Equations \ref{eq:motortorque}, \ref{eq:noslipcondition}, and \ref{eq:turrettorque}, and the selected motor specifications are in Table \ref{tab:motorselect}. Comparing the torque capabilities of the motor to the results in Fig. \ref{fig:numlinklimit} shows that CLIMR can accommodate a wide range of column diameters, but there is room for improvement. 

\section{VERIFICATION}
\label{sec:verification}

\begin{figure}
\vspace{0.1 in}
    \centering
    \includegraphics[width = 0.9 \linewidth]{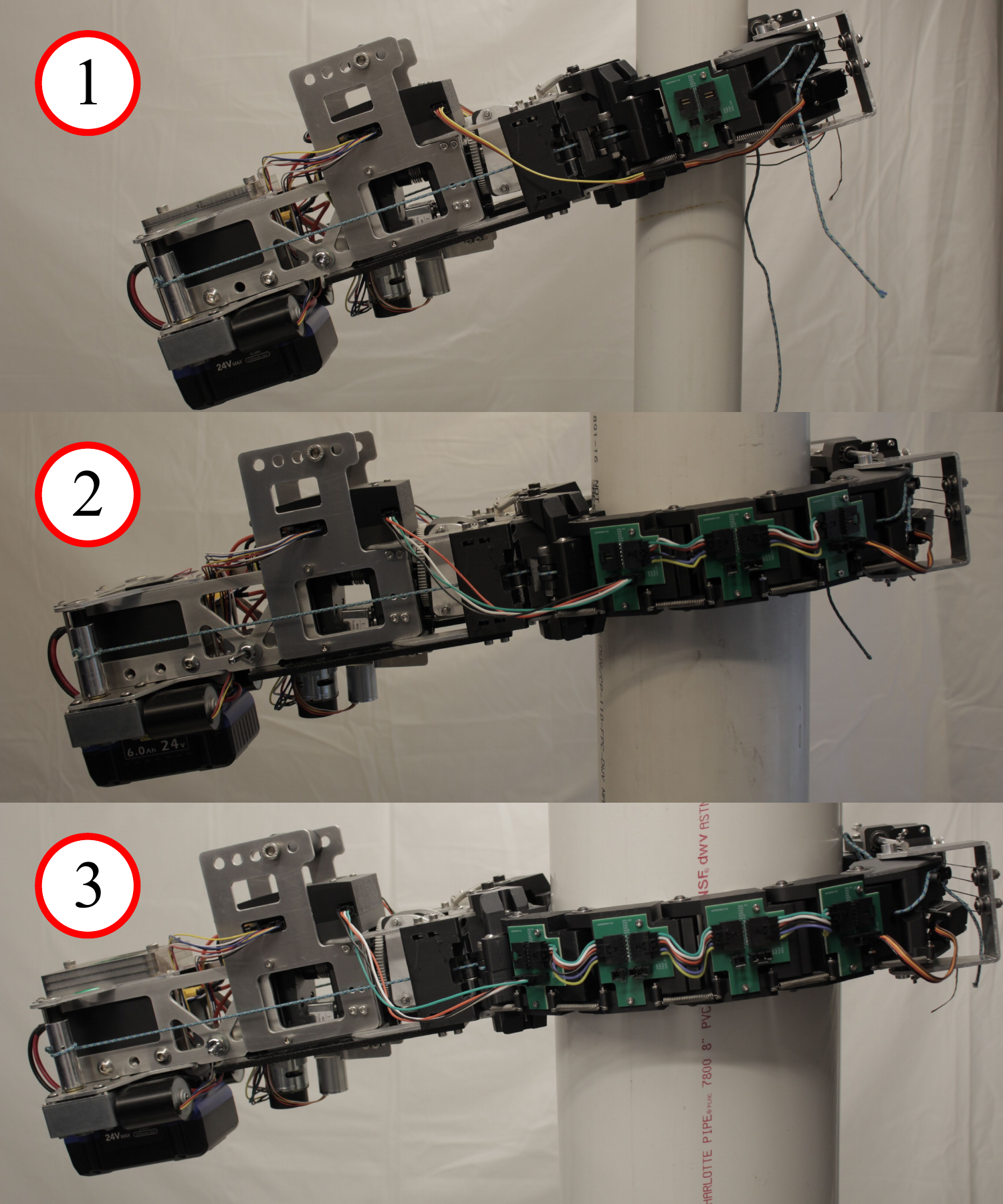}
    \caption{Self locking tests on columns with diameters of 90 mm (1), 170 mm (2), and 220 mm (3). Note the variation in added tail weights. The tail without weights sufficed for self-locking, but varying the masses tested how robust the robot's climbing ability was to errors in mass distribution.}
    \label{fig:experimentselflocking}
\end{figure}
The self-locking capability of the robot was tested on PVC pipes of different diameters: 90 mm, 170 mm, and 220 mm. The required number of links was found using Fig. \ref{fig:numlinklimit}, and the arm length was adjusted accordingly. Fig. \ref{fig:experimentselflocking} shows the robot demonstrating self-locking on the three different pipes, illustrating that the selected $d_1$ is sufficient. As seen in the supplemental video, successful rotation about the column, helical movement up the column, and autonomous grasping were also achieved.

\begin{figure}
    \centering
    \includegraphics[width = 0.99 \linewidth]{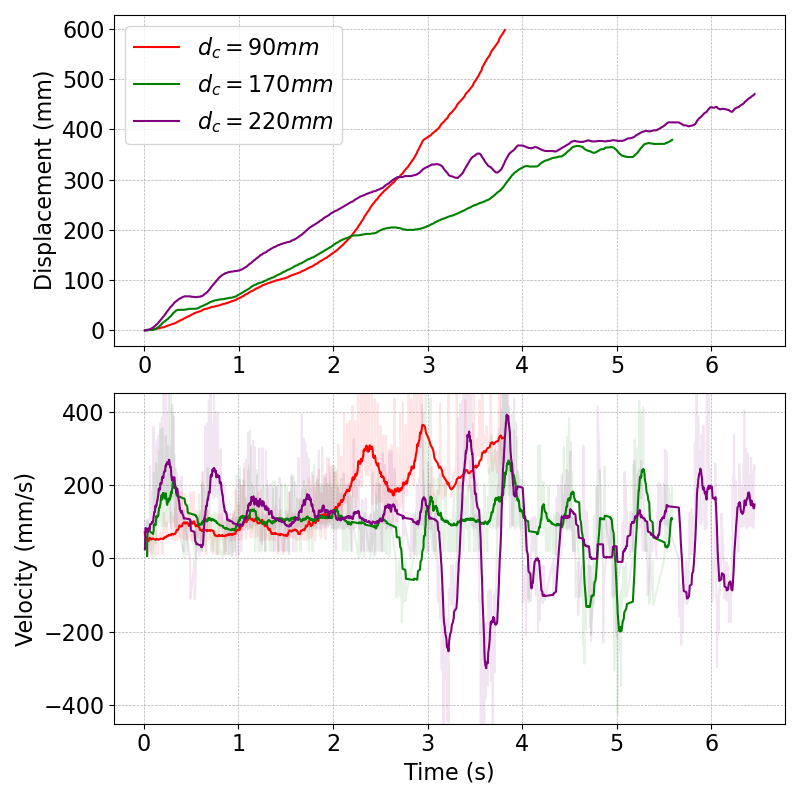}
    \caption{Position and velocity measurements for different column diameters. In the lower subplot, the actual velocity measurements are shown in faded colors, while the solid colored lines indicate the velocity measurements smoothed out with a moving average filter (window = 30).}
    \label{fig:experimentvelocity}
\end{figure}
The position and velocity of CLIMR was also measured using a TinkerForge Distance IR Bricklet 2.0 mounted at the base of the pipe under the robot's tail. The results when the drive wheel is given a position command of 20 rad can be seen in Fig. \ref{fig:experimentvelocity}. The plot shows that wheel slipping occurs, which was to be expected with the polished slippery surface of the PVC pipes. Qualitatively, the pipes varied in slipperiness---with 170 mm being the most, followed by 220 mm and then 90 mm---and the variation in wheel slip between pipes highlights this as well as the difficulty in keeping the climbing surface consistent between different pipe tests. Furthermore, more wheel slip occurred at larger $d_c$, which is most likely due to the extra mass of the extra links shifting the center of mass closer to the column axis.

Regardless, Fig. \ref{fig:experimentvelocity} demonstrates successful climbing of different diameter columns in what is considered the ``worst-case scenario," where friction between the pipe and the wheel is low ($u_w=0.7$). The average velocities of CLIMR on the 90, 170, and 220 mm pipes were 144.49, 102.01, and 100.07 mm/s, respectively. However, the actual velocities were most likely higher than that, as the oscillations of the tail were picked up as negative velocity measurements even though the overall movement of the robot was upwards. This is better perceived in the supplemental video. Nevertheless, these velocities are higher than or comparable to past wheeled climbing robots such as \cite{liu2021, gui2017, fauroux2010}.

\section{CONCLUSIONS}

This paper presents a novel modular hybrid climbing robot and introduces a structured method to universally evaluate the performance of similar designs. For the presented design, future work to increase the maximum column diameter limit includes manufacturing custom ball transfers to reduce linkage weight and redesigning the robot with a more powerful motor farther out on the tail to replace non-functional counterweight with functional motor windings. Increasing the PCB copper thickness would also increase the motor current draw limit and thus the allowable drive torque. Lastly, testing CLIMR on different surfaces would expand on its capabilities in non-ideal scenarios.   









\bibliographystyle{IEEEtran}
\bibliography{zoteroreferences.bib, ryanfull.bib}

\end{document}